\pdfoutput=1

\documentclass[11pt]{article}

\usepackage[final]{acl}

\usepackage{times}
\usepackage{latexsym}
\usepackage{booktabs}
\usepackage[T1]{fontenc}
\usepackage{amsmath}
\usepackage{multirow}

\usepackage[utf8]{inputenc}

\usepackage{microtype}

\usepackage{inconsolata}

\usepackage{graphicx}
\usepackage{hyperref} 

\title{Recover-LoRA: Data-Free Accuracy Recovery of Degraded Language Models via Low-Rank Adaptation}

\author{Devleena Das, Rajeev Patwari, Ashish Sirasao \\
       Advanced Micro Devices, Inc. (AMD) \\ \{devleena.das, rajeev.patwari, ashish.sirasao\}@amd.com \\ }

\begin{document}
\maketitle
\begin{abstract}
Inference optimizations such as quantization, pruning, format and datatype conversion, model export, and serialization can lead to functional degradations in language model task performance. While most efforts on performance recovery for deployment focus on robust quantization techniques, we focus on recovering model accuracies from any sources that degrade model weights, such as improper model serialization. In this work, we propose Recover-LoRA, a lightweight and dataset agnostic method to recover accuracy in degraded models. Recover-LoRA uses synthetic data and logit distillation to learn LoRA adapters on selective layers that facilitate aligning the degraded model to its full precision model. We investigate the utility of Recover-LoRA across a diverse set of small language models (SLMs), including models with varying attention architectures, multi-head attention (MHA) and group-query attention (GQA), as well as several evaluation datasets. Our results show that Recover-LoRA recovers model accuracies by 5-17\% on MHA and GQA SLMs.
\end{abstract}

\section{Introduction}
Small language models (SLMs), typically under 5B parameters, have shown strong capabilities on downstream tasks while offering a smaller memory and compute footprint compared to their larger language model counter parts (i.e. Phi3.5-mini, Llama3.2 1B, etc.)\cite{lu2024small}. These smaller models have become of popular interest for edge deployment where compute, memory and latency are critical bottlenecks \footnote{\url{https://blogs.windows.com/windowsexperience/2024/12/06/phi-silica-small-but-mighty-on-device-slm/}}
. However, for edge deployment, language models often undergo further optimization or conversion steps that can inadvertently introduce accuracy degradation due to structural inconsistencies or weight corruptions. For example, accuracy loss can stem from quantization \cite{zhu2024survey}, sparsity \cite{zafrir2021prune}, improperly saving or loading model states, custom layers, or format drifts when transferring among tool chains (eg. Pytorch to ONNX). These scenarios may result in packaged models that are structurally sound, where shapes and architectures are preserved, but downstream task performance significantly varies from the original model. 

Among these sources of error, quantization is one of the most popular studied, given its importance  for reducing inference latency and memory footprint \cite{zhu2024survey}. Post-Training Quantization (PTQ) methods such as AWQ \cite{lin2024awq} convert weights to lower precision without retraining, while Quantize-Aware Training (QAT) retrains the model with simulated quantization noise \cite{lang2024comprehensive}. Most recently, LLM-QAT \cite{liu2023llm} has shown that synthetic data, instead of labelled data, can be used to perform QAT for LLama models. 

While our work is inspired by QAT and LLM QAT to preserve model accuracy, we focus on recovering accuracy from more  sources of errors, outside of quantization, that can occur in deployment settings, leading to corrupted model weights. Additionally, we consider the practical constraints within industry settings such as scarce labeled data or proprietary data, and minimizing retraining of large models. To this end, we explore the following: \textit{how can we recover model accuracy loss without requiring full, model training and utilize synthetic data?} 



In this work, we introduce \textbf{Recover-LoRA} a lightweight, dataset agnostic approach to recovering accuracy from functionally degraded models where the model weights have undergone silent corruption. Recover-LoRA leverages synthetic data, inspired by LLM QAT \cite{liu2023llm}, to learn low-rank matrices (LoRA adapters \cite{hu2022lora}) that align the corrupted model with its full-precision, reference language model via logit distillation \cite{gou2021knowledge}. In this manner, Recover-LoRA provides a parameter-efficient approach to accuracy recovery, while providing data independence through synthetic data. While LoRA \cite{hu2022lora} is a common lightweight finetuning approach, it is traditionally applied to task adaptation with labeled datasets. To the best of our knowledge, Recover-LoRA is the first to consider the feasibility of LoRA adapters in recovering degraded model accuracy.


We study the efficacy of Recover-LoRA across four different SLM architectures, including multi-head and group-query attention models (MHA, GQA), using functionally degraded models derived from improper model weight serialization, and evaluate on seven different datasets. Our work contributes the following:
\begin{enumerate}
    \item We introduce Recover-LoRA, to the best of our knowledge, as the first approach to recover lost model accuracy in dedgraded models. Recover-LoRA provides a lightweight and data-flexible method to restore model performance by learning LoRA adapters with logit distillation, and using synthetic data.
    \item We demonstrate that Recover-LoRA effectively improves model accuracy in MHA and GQA style models. We show an average accuracy recovery ranging from 5\% to 17\%, surpassing the recovery capabilities of LLM QAT \cite{liu2023llm} across all tested models, and surpassing dataset-specific LoRA finetuning on three out of the four tested models.
\end{enumerate}


\section{Related Work}
\subsection{Sources of LLM Accuracy Degradation}
LLM accuracy degradation can occur due to several factors including quantization \cite{zhu2024survey}, sparsity \cite{zafrir2021prune}, framework conversion \cite{louloudakis2023fix}, datatype conversion \cite{rouhani2023microscaling}, etc. Below we describe key sources of error related to our work.

Recently, Jalal et al. \cite{jajal2023analysis} highlight the common failure points in ONNX conversion, whereas FetaFix \cite{louloudakis2023fix} proposes an automated approach to detect and repair models conversions between deep learning frameworks. Similarly, 
state of the art accelerators support fast microscaling formats for inference \cite{rouhani2023microscaling} like $MXFP6$, $MXFP8$, and $MXINT8$. Post-training model conversion to such data types may degrade the quality of the LLM specific to the application.
Additionally, sparsity techniques that aim to prune model weights can also lead to degraded model performance \cite{zafrir2021prune}. These scenarios indicate that conversions for deployment can lead to degraded model performance, highlighting a need for accuracy recovery. Recover-LoRA aims to provide a lightweight method for recovering degraded model performance specifically considering silent failures from model weight serialization.

\paragraph{Quantization for LLMs}
Quantize-Aware Training (QAT) techniques are widely adopted to reduce impact of quantization specific accuracy degradation. For example, DL-QAT combines group-wise scaling with LoRA based updates to further improve QAT efficiency \cite{ke2025dl}. While most QAT approaches use labeled data, LLM QAT \cite{liu2023llm} shows the utility of synthetic data for QAT. LLM QAT \cite{liu2023llm} generates synthetic training data from a full-precision LLaMA 7B model and uses knowledge distillation to train several quantized LLaMA models. 

Our work is inspired by the usage of synthetic data in LLM QAT \cite{liu2023llm}, but we focus on error stemming from functional degradation  not limited to quantization. Specifically, we use synthetically generated data from a pretrained SLM to align the degraded model. Also, unlike LLM QAT, we limit model updates to solely LoRA adapters and enable a more efficient method to accuracy recovery in degraded models. 

\subsection{Pruning and Recovery Techniques}
Recent work has also explored compressing LLMs via pruning and recovering performance post-compression. For example, Minitron \cite{sreenivas2024llm} introduces a multi-stage pipeline involving teacher correction using labeled datasets, followed by structured pruning and knowledge distillation to produce competitive, optimized models. Additionally, Thangarasa et al. \cite{thangarasa2024self} propose a self-data distillation approach to recover accuracy in pruned models. Specifically, the authors utilize existing fine-tuning datasets and access to a full teacher model to generate distilled outputs which are then used for accuracy recovery. Both Thangarasa et al. \cite{thangarasa2024self} and Minitron \cite{sreenivas2024llm} rely on access to labelled datasets, whereas our Recover-LoRA operates in a data-free setting without any reliance on labelled data. 

\vspace{-0.1cm}

\subsection{Parameter-Efficient Fine Tuning (PEFT)}
PEFT updates a smaller set of model parameters, compared to all model parameters, to improve computational efficiency during the training process \cite{ding2023parameter}. A common PEFT approach is LoRA \cite{hu2022lora} in which low-rank matrices, known as adapters, are learned during finetuning. 
In some PEFT methods the pretrained model is quantized while the LoRA adapters are trained in higher precision. For example, in QLoRA \cite{dettmers2023qlora} the pretrained model is quantized to NF4, whereas in QA-LoRA \cite{xu2023qa}, it is quantized to INT4. LoRA is primarily motivated to improve training efficiency for task-specific adaptation \cite{mao2025survey}.  Our work studies the use of LoRA beyond task adaptation and considers the utility of LoRA as a lightweight approach to recover functionally degraded model accuracy due to improper serialization.


\begin{figure*}[t!]
\centering
\includegraphics[width=13cm]{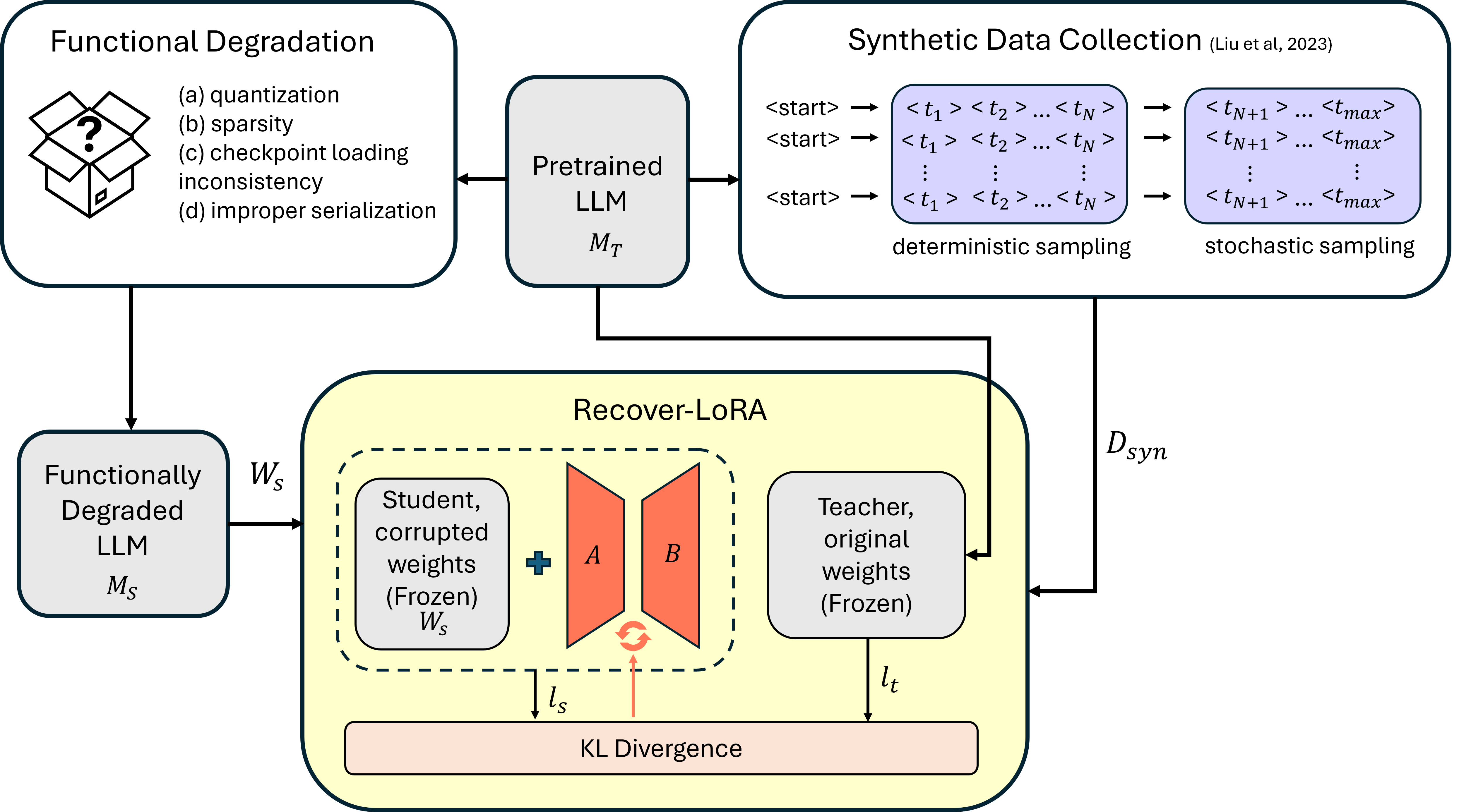}
\caption{Recover-LoRA recovers model accuracy by leveraging logit distillation to align an improper weight serialized model, $M_S$, to its pretrained LLM, $M_{T}$, by learning LoRA adapters, $A$ and $B$, with a synthetically generated dataset $D_{syn}$. }
\label{fig:overview}
\end{figure*}

\section{Background}
\label{sec:background}
\subsection{LoRA}
LoRA \cite{hu2022lora} is a PEFT approach in which low-rank matrices, known as adapters, are trained and added to the pretrained model's frozen weights. Let $W \in R^{d\times k}$ represent the pretrained weights where $d$ and $k$ define the output and input dimensions. LoRA then defines two trainable matrices $A \in R^{r \times k}$ and $B \in R^{d \times r}$ where $r << (d,k)$ represents the rank of the LoRA matrices. During finetuning, $W$ is frozen and only $A$ and $B$ are updated. The LoRA output, $Y$,  for a given layer is then represented as:
\vspace{-0.3cm}
\begin{equation}
\vspace{-0.3cm}
\label{eq:lora}
    Y=WX + \alpha BAX
\end{equation}
where $X$ represents the input activation, and $\alpha$ represents a scaling factor that controls the contribution of LoRA adapters on $Y$.

\subsection{Knowledge Distillation}
Knowledge distillation aligns the outputs of smaller student model with the outputs of a larger teacher model \cite{hinton2015distilling, gou2021knowledge, liu2023llm}. Let $M_{T}$ represent the teacher model and $M_{S}$ represent the student model. During training, $M_{S}$ is optimized by minimizing the Kullback-Leibler (KL) divergence between the predicted probabilities of $M_{S}$, $p_{s}$, and the soft-target probabilities of $M_{T}$, $p_{t}$ \cite{sanh2019distilbert}. The loss function is defined as:
\vspace{-0.3cm}
\begin{equation}
\vspace{-0.3cm}
    L_{KD} = KL(p_{t}||p_{s}) = \sum_{i}p^{i}_tlog\dfrac{p^i_t}{p^i_s}
\label{eq:KD}
\end{equation}

\subsection{Functionally Degraded Models}

Pretrained model accuracy degradation can be caused by many factors such as improper serialization, quantization, sparsity, and ONNX export, to name a few. We simulate improper weight serialization by introducing minor perturbations to the attributes of $torch.nn.Linear$ for K and V projection layers and save the pretrained model using the HuggingFace $save\_pretrained()$ API. The result is noisy saved model weights that deviate from the original weights.

\section{Methodology}
\label{sec:method}
Figure \ref{fig:overview} provides an overview of our approach, Recover-LoRA, which aims to recover accuracy lost in functionally degraded models in a lightweight and dataset agnostic manner. Specifically, Recover-LoRA takes as input, 
$M_{S}$, the degraded model, and $M_T$, the pretrained, full precision model. Note, Recover-LoRA does not require any knowledge of the type of functional degradation, and instead only requires access to $M_{S}$ and $M_{T}$.
In our application, we assume $M_{S}$ has degraded performance due to improper weight serialization. Recover-LoRA then learns key LoRA adapters to align the adapter weights to the pretrained language model's weights via logit distillation. Training only LoRA adapters makes Recover-LoRA a lightweight approach to recover error from weight-corrupted models. 
Additionally, the dataset $D_{syn}$ utilized for training is not a
labeled dataset. Instead, $D_{syn}$ represents synthetic data generated using the hybrid sampling method outlined in LLM QAT \cite{liu2023llm}, making the Recover-LoRA training process data-flexible.

\begin{table}[t!]
\centering
\resizebox{0.4\textwidth}{!}{%
\begin{tabular}{| l | c |}
\hline
\textbf{Model} & \textbf{L2 Norm} \\
\hline
AMD-Olmo-SFT 1B & 44.06 \\
Llama3.2 1B     & 52.97 \\
Gemma2 2B       & 35.94 \\
DeepSeekR1 Distill Qwen 1.5B & 40.69 \\
\hline
\end{tabular}
}
\caption{L2 norm difference between original and perturbed weights, indicating model degradation.}
\label{table:l2-norm}
\vspace{-0.5cm}
\end{table}

\subsection{Functionally Degraded LLM}
We insert error into the LLM by introducing minor perturbations to the weight attributes of torch.nn.Linear for K and V projections and saving the model with HuggingFace's $save\_pretained()$. Our functionally degraded LLM simulates incorrect weight serialization. In Table \ref{table:l2-norm}, we show the L2 norm difference between the original weights and perturbed weights for the first K projection layer of several models to indicate the random noise that is introduced. 

\subsection{Synthetic Data Collection}
The LoRA adapters in Recover-LoRA are trained with synthetic data generated through a hybrid sampling strategy outlined in LLM QAT \cite{liu2023llm}. Specifically, a pretrained language model deterministically generates the first 3-5 tokens, and stochastically generates the remaining tokens, balancing stability and diversity. While LLM QAT \cite{liu2023llm} studies hybrid sampling in a QAT setting, we explore the utility of synthetic data in broader functionally degraded model settings. Details on the hybrid sampling hyperparameters used in Recover-LoRA are provided in Appendix \ref{sec:syn-data}.

\subsection{Recover-LoRA}
Recover-LoRA aims to improve the accuracy of $M_{S}$ by learning a set of lightweight LoRA adapters, $A$ and $B$, using logit distillation. From Equation \ref{eq:lora}, the weight matrix $W$ in Recover-LoRA is represented as $W_{s}$,
the frozen, corrupted weight matrix from improper weight serialization (see Sec. \ref{sec:background}). Adapters $A$ and $B$ are optimized by minimizing the KL divergence between the predicted logit distributions of $M_{S}$ and $M_{T}$. Following Equation \ref{eq:KD}, $p_{s}$ and $p_{t}$ represent student and teacher logits, $l_{s}$ and $l_{t}$. The logits are derived as $l_{t} = softmax(M_{T}(x))$ and $l_{s} = softmax(M_{S}(x))$, where $x$ is a training sample.

While LoRA is a PEFT method for task adaptation, we examine a new use case of LoRA adapters, focusing on restoring model accuracy in degraded models due to corrupted weights. In Section \ref{sec:results}, we demonstrate the success of Recover-LoRA in recovering degraded model accuracies in both a parameter and a data-efficient manner.

\begin{table*}[t!]
\resizebox{\textwidth}{!}{%
\renewcommand{\arraystretch}{1.4} 
\begin{tabular}{c| c | p{2.0cm} | cccccccc | c }\toprule
        & \centering \textbf{Method} & \centering \textbf{LoRA Adapters} & \textbf{HellaSwag} & \textbf{MMLU Avg.} &\textbf{Arc C} & \textbf{WinoGrande} & \textbf{PiQA} & \textbf{OpenbookQA} & \textbf{BoolQ} & \textbf{Avg} & \textbf{AR\%} \\
    \midrule
       \multirow{5}{*}{\rotatebox[origin=c]{90}{\parbox{3cm}{\centering AMD OLMO SFT 1B}}} & $M_T$ & \centering -- & 28.38 & 33.42  & 24.32  & 51.3 & 61.43 & 18.2 & 55.44 &  38.93 & -- \\
        & $M_{S}$  & \centering -- & 25.42 & 31.41  & 21.84  & 50.28 & 53.05 & 15.4 & 43.85 & 34.46 & -- \\
        & Recover-LORA & \centering K,V & 25.54 & 17.96  & 20.56 & 50.83 & 53.97 & 15.6 & 62.17 & 35.23 & \textbf{17.24} \\
        & LLM-QAT* & \centering -- & 27.69  & 27.71  & 21.08 & 50.83 & 55.88 & 15.2 & 39.63 & 34.00 & -10.34 \\
        & SFT LORA & \centering K,V & 24.48 & 31.84  & 23.29 & 48.46 & 52.94 & 18.8 & 46.21  & 35.15 & 15.29 \\

        \midrule \midrule
        
        \multirow{5}{*}{\rotatebox[origin=c]{90}{\centering LLAMA3.2 1B}} & $M_T$ & \centering -- & 47.74 & 41.77 & 31.48 & 60.93 & 74.27 & 26.8 & 63.73 & 49.53 & -- \\
        & $M_{S}$ & \centering -- & 25.51 & 28.58  & 21.93  & 50.83 & 54.3 & 17.00 & 37.89 & 33.72 & -- \\
        & Recover-LORA & \centering ATTN, MLP & 25.69 & 32.06 & 21.33 & 50.12 & 53.54 & 16.8 & 51.31 & 35.84 & \textbf{13.38}\\
        & LLM-QAT* & \centering -- & 25.72  & 17.96  & 20.73 & 48.78 & 53.75 & 14.6 & 38.17 & 31.39 & -14.75\\
        & SFT LORA & \centering ATTN, MLP & 25.59& 23.66  & 22.01 & 49.41 & 51.85 & 18.6 & 51.99  & 34.73 & 6.39\\

        \midrule \midrule

        \multirow{5}{*}{\rotatebox[origin=c]{90}{\centering GEMMA2 2B}} & $M_T$  & \centering-- & 54.99 & 56.75  & 46.84 & 68.75 & 78.67 & 31.4 & 73.58 & 58.71 & --\\
        & $M_{S}$ & \centering -- & 25.92 & 22.81  & 20.73  & 50.9 & 53.05 & 17.6 & 45.93  & 33.85 & -- \\
        & Recover-LORA & \centering ATTN, MLP & 25.98 & 17.76 & 20.73 & 50.51 & 52.72 & 14.6 & 41.68 & 31.99 & -7.45\\
        & LLM-QAT* & \centering -- & 26.26 & 24.61  & 18.26 & 48.38 & 54.9 & 11.4 & 28/13 & 31.71 & -8.62\\
        & SFT LORA & \centering K,V & 35.21 & 25.04  & 24.23 & 52.09 & 67.37 & 21.00 & 61.22  & 40.88 & \textbf{28.28} \\

        \midrule \midrule

        \multirow{5}{*}{\rotatebox[origin=c]{90}{\parbox{3cm}{\centering DeepSeek R1 Distill Qwen 1.5B}}} & $M_T$ & \centering -- & 36.39 & 44.9 & 34.47 & 55.88 & 65.29  & 20.2 & 68.01 & 46.45 &  -- \\
        & $M_{S}$ & \centering --  & 25.93 & 20.64 & 21.08  & 48.54 & 52.83 & 16.4 & 59.6 & 35.00 & -- \\
        & Recover-LORA  &\centering  K,V & 26.52 & 22.85  & 18.52 & 49.72 & 54.79 & 15.2 & 61.38 & 35.6 & \textbf{4.95}\\
        & LLM QAT* & \centering -- &  26.04  & 19.05  & 20.31 & 50.36 & 54.35 & 14.00 & 59.72 & 34.93 & -1.49\\
        & SFT LORA & \centering K,V &  27.75 & 27.29  & 20.73 & 49.8 & 57.24  & 14.4 & 48.44  & 35.09 & 0.79\\
    \bottomrule
\end{tabular}}
\caption{Average accuracy recovery percentage (AR\%) comparisons for all recovery techniques and model comparisons. Note, $M_{T}$ represents the pretrained SLM, and $M_{S}$ is the degraded model. }
\label{table:overall-AR-percents}
\end{table*}

\section{Experiments}
\label{sec:experiment}
We detail the experimental setup used to evaluate Recover-LoRA. We finetune using AMD MI300X GPUs and describe all hyperparameters in Appendix \ref{sec:finetuning-hyperparams}. 

\subsection{Baselines}

\paragraph{LLM QAT*} Our primary baseline is LLM QAT \cite{liu2023llm}, which uses synthetic data generated for QAT, via knowledge distillation, to produce quantized LLaMA models. We compare with an adaptation of LLM QAT, \textbf{LLM QAT*}, where we do not perform the original QAT process of simulating quantization effects in training. Instead, LLM QAT\* takes an improper serialized model and performs logit distillation on all the model parameters to align the degraded model to its pretrained, teacher model using synthetic data. 

\paragraph{SFT LoRA}
We also compare with the traditional supervised finetuning (SFT) LoRA approach which uses good quality labeled datasets to finetune the degraded model, via a cross-entropy loss. Specifically, we leverage the OpenHeremes-2.5, WebInstructSub and Code-Feedback datasets for finetuning, which prior work\footnote{\label{note1} \url{https://huggingface.co/amd/AMD-OLMo-1B-SFT}} has established appropriate for seeing improvements on our designated evaluation tasks. By using these labeled datasets, we measure the effect of using synthetic data and logit distillation compared to high-quality labeled data for model accuracy recovery.

\subsection{Evaluation Datasets and Models}
\paragraph{Evaluation Datasets}
We evaluate across seven different datasets. Specifically, we evaluate commonsense reasoning with PiQA \cite{bisk2020piqa}, OpenBookQA \cite{mihaylov2018can}, WinoGrande \cite{sakaguchi2021winogrande}, BoolQ \cite{clark2019boolq}, HellaSwag \cite{zellers2019hellaswag}, Arc Challenge (ARC C) \cite{clark2018think} and multi-task factual knowledge with three randomly selected subsets of MMLU \cite{hendrycks2020measuring}, MMLU Philosophy, Management, Astronomy. 

\paragraph{Evaluation Models}
We apply Recover-LoRA to four SLMs to assess its generalizability in recovering degraded model accuracy: \textbf{Gemma2 2B} \cite{team2024gemma}, \textbf{Llama3.2 1B} \cite{grattafiori2024llama}, \textbf{DeepSeek-R1-Distill-Qwen 1.5B} \cite{guo2025deepseek} and \textbf{AMD-Olmo-SFT 1B} \textsuperscript{\ref{note1}}. These models represent a diverse set of architectures with different attention mechanisms (see Appendix \ref{sec:eval-models} for more details).

\subsection{Metrics}
We define \textbf{Accuracy Recovery Percentage (AR\%)}, to measure the efficacy of Recover-LoRA: 
\begin{equation}
    AR\%= \dfrac{(E^*_{S} - E_{S})}{|E_{S} - E_T|}*100
\end{equation} where, 
$E_{S}$ and $E_{T}$ represent evaluation scores of the degraded ($M_{S}$) and full-precision ($M_{T}$) SLMs, respectively. Additionally, $E^*_{S}$ represents the evaluation scores of the functionally degraded model, after applying one of the three error recovery techniques, Recover-LoRA, LLM QAT* or SFT LORA. The metric computes how much accuracy is recovered from the degraded model, via a given recovery technique. If $AR\% = 100$, all error is recovered and $E_{S} = E_T$, and if $AR\% = 0$, no error is recovered. If $AR\% < 0$, the recovery technique worsened the error. 

\section{Results}
\label{sec:results}
\subsection{Accuracy Recovery from Recover-LoRA}
Table \ref{table:overall-AR-percents} shows the average AR\% of each recovery method. Overall, we observe that Recover-LORA outperforms LLM QAT* and SFT LoRA across three of the four models: AMD-OLMO-SFT 1B, LLaMA3.2 1B and DeepSeek-R1-Distill-Qwen 1.5B. Interestingly, we see negative AR\% from LLM QAT* in all models, showing that LLM QAT* worsens the functional degradation error. We hypothesize that this is due to LLM QAT* updating all model parameters, which may cause overfitting, whereas Recover-LoRA updates a smaller fraction of model parameters. Also, we observe that SFT LoRA performs best for GEMMA2 2B, suggesting that training with synthetic data may be ineffective in some models. We hypothesize that GEMMA2 2B's architecture may be more sensitive to distributional mismatches between the synthetically data and its pretraining data, or more training epochs may be needed for Recover-LoRA to be effective.


\begin{figure}[t!]
\centering
  \vspace{-0.4cm}
  \includegraphics[width=0.42\textwidth]{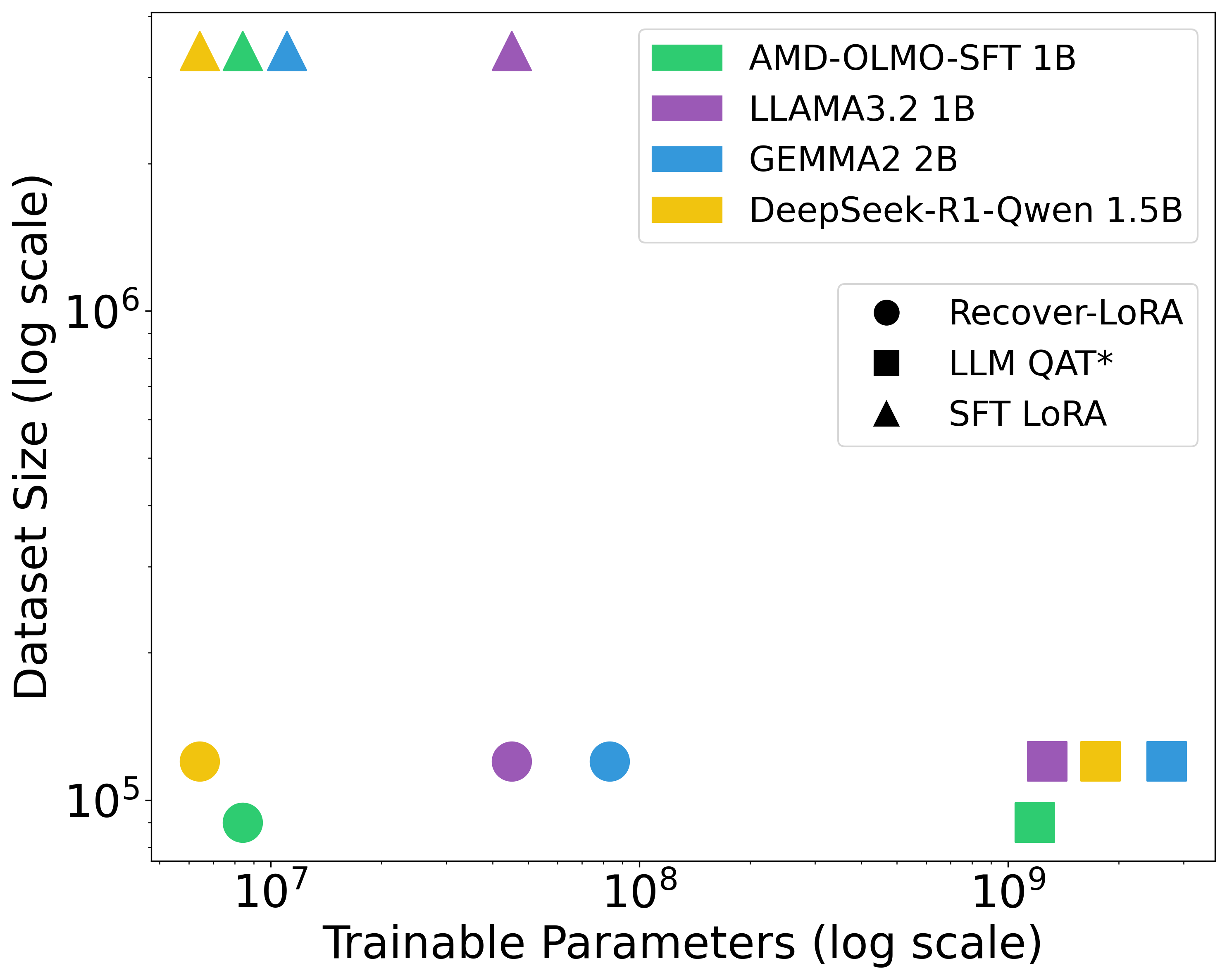}
  \caption{Trainable parameters and dataset size comparisons for all recovery methods, showing the parameter and data efficiency of Recover-LoRA.}
  \label{fig:params-dataset-size}
\end{figure}

\subsection{Parameter \& Data Efficiency}
Figure \ref{fig:params-dataset-size} shows the amount of training data used by each recovery method, for each model, to achieve the AR\% in Table \ref{table:overall-AR-percents}. Specifically, Recover-LoRA and LLM QAT* utilize 90k synthetic samples for AMD-OLMO-SFT, and 120k samples for all other models. In contrast, SFT LoRA utilizes a fixed, labeled dataset of 3M samples, previously selected by prior work to improve commonsense reasoning and multi-knowledge task performance (see Sec. \ref{sec:experiment}). Overall, we observe that Recover-LoRA achieves high AR\% with less trainable parameters than LLM QAT* and less data than SFT-LoRA.

\subsection{Synthetic Datasets in Recover-LoRA}
\begin{figure}[t!]
\centering
  \includegraphics[width=0.42\textwidth]{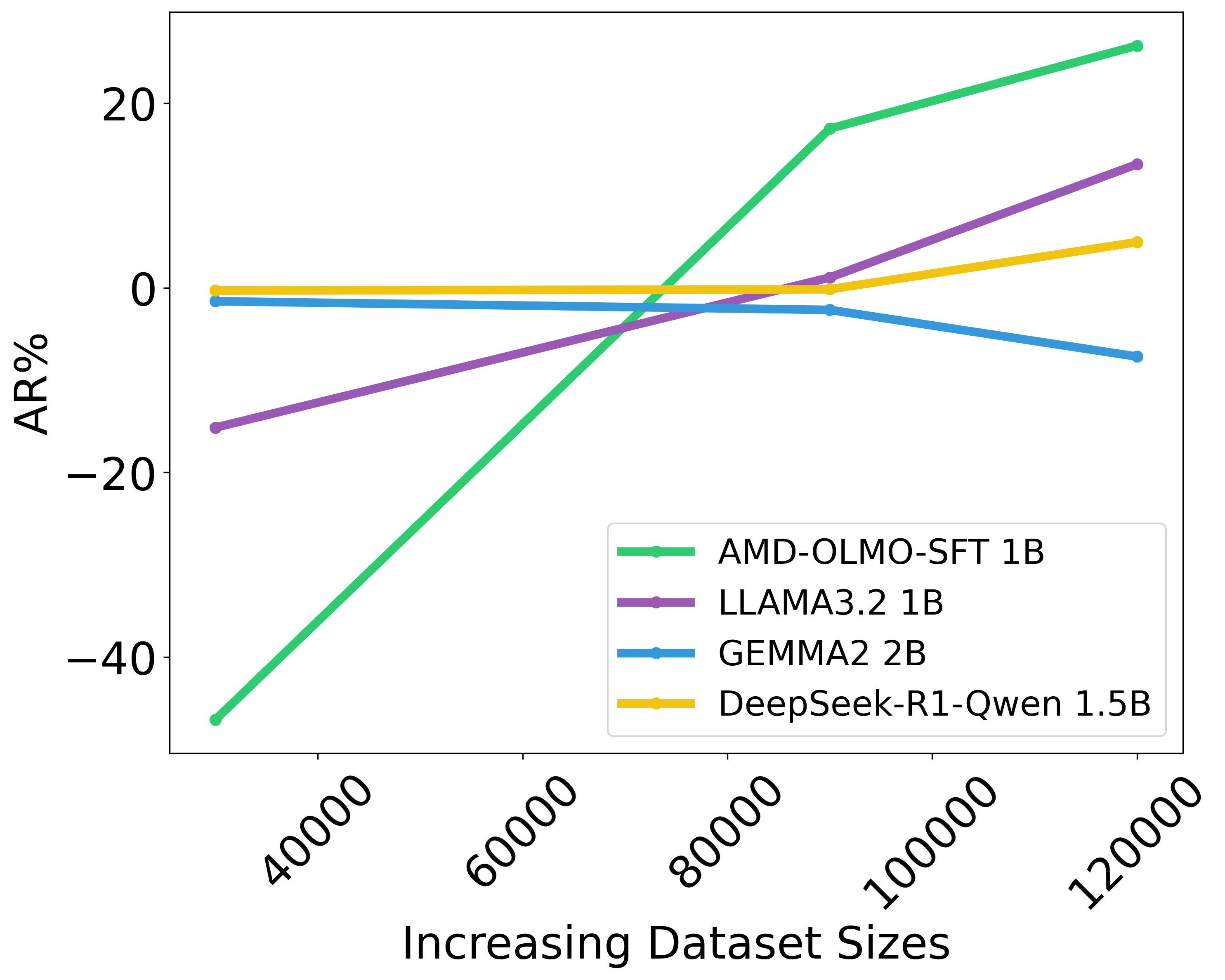}
  \caption{Progression of AR\% with increasing dataset size, showing a minimum of 90k synthetic data samples are needed positive AR\% in three models.}
  \label{fig:ar-dataset}
\end{figure}

The synthetic datasets utilized in Recover-LoRA enable a data-independent functional model degradation recovery method where good quality labeled data are not needed for training. Figure \ref{fig:ar-dataset} shows that Recover-LoRA uses a minimum of 90k samples for positive AR\% in three of the four models, with more data yielding higher AR\%. We hypothesize that applying Recover-LoRA to larger models will require more synthetic data. But more importantly, we show the flexibility of using synthetic data in Recover-LoRA, and that, depending on the application, such synthetic data can be readily generated and utilized.

\subsection{Practical Development \& Usage}
While Recover-LoRA demonstrates strong accuracy recovery and efficiency across AMD-OLMO-SFT 1B, LLaMA3.2 1B and DeepSeek-R1-Distill-Qwen 1.5B, below we present several considerations for practical deployment. First, adapter placement can significantly impact recovery performance. As shown in Table \ref{table:overall-AR-percents}, some models benefit from LoRA adapters on K and V projection layers, while other models benefit from adapters on all attention and MLP layers. Therefore, for practical deployment, a systematic search is necessary for identifying optimal adapter configurations per model. Additionally, the choice of model used for synthetic data generation influences recovery effectiveness. Specifically, Recover-LoRA works best when synthetic data is generated from a pretrained SLM that shares the same vocabulary and tokenizer as the degraded SLM. We provide these details in Appendix \ref{sec:syn-data}. Also, a practical challenge posed with Recover-LoRA is diagnosing scenarios of limited recovery. Such limitations may stem from factors including suboptimal LoRA adapter configuration, insufficient synthetic data, or from architectural constraints of the degraded model itself. In the latter case, Recover-LoRA may be fundamentally limited in its ability to restore performance. Understanding these distinctions is crucial for maximizing Recover-LoRA's effectivenss and ensuring its practical usability across models. 
 

\section{Conclusion}
We introduce Recover-LoRA, a lightweight, dataset agnostic method to recover degraded model accuracy. Recover-LoRA leverages synthetic data to train LoRA adapters by using logit distillation to align a functionally degraded model with its pretrained SLM. Recover-LoRA does not require knowledge of the type of functional degradation. In this manner, Recover-LoRA provides a practical solution for recovering model degradation without requiring full model retraining or access to labeled data. 
Our results show the efficacy of Recover-LoRA in improving degraded model accuracies by 5-17\%, while showcasing its parameter and data efficiency, highlighting its use case for real-world deployment.
\paragraph{Limitations} Our results show Recover-LoRA to be effective for some MHA and GQA architectures. We also highlight that LoRA adapters are model-dependent in Recover-LoRA. Future work should investigate expanding the capabilities of Recover-LoRA to MQA \cite{shazeer1911fast}, MLA \cite{liu2024deepseek} architectures; and how to automatically select the minimal set of LoRA adapters needed per model architectures using methods like Neural Architecture Search (NAS) \cite{ren2021comprehensive}. Additionally, future work should examine the applicability of Recover-LoRA on larger language models ranging between 7B-13B. Lastly, more experiments are needed to study the generalizability of Recover-LoRA in recovering accuracy from other sources of accuracy degradation such as quantization and pruning.


\bibliography{custom}

\appendix

\section{Synthetic Data Collection Hyperparameters}
\label{sec:syn-data}
The hybrid sampling strategy in LLM QAT \cite{liu2023llm} generates the first few tokens greedily and the remaining tokens stochastically. In our usage of hybrid sampling, we set the number of greedily generated tokens to 5, and allow stochastic generation up to a max sequence length of 2048. The pretrained SLM is selected such that its vocabulary and tokenizer match with the degraded SLM to allow for meaningful training in Recover-LoRA. For example, our degraded Llama3.2 1B model utilized the pretrained Llama3.2 1B model for data generation. Our degraded Deepseek-R1-Distill-Qwen 1.5B model utilized the pretrained Llama3.2 1B model for data generation, since the Deepseek-R1-Distill-Qwen models utilize the LlamaFastTokenizer. Similarly, our degraded Gemma2 2B model utilized the pretrained Gemma2 2B model for data generation, and our degraded AMD-OLMO-SFT 1B model utilized the pretrained AMD-OLMO-SFT 1B model for data generation. Section \ref{sec:results} demonstrates the utility of hybrid sampling in recovering degraded model accuracy and details our ablation studies on the amount of synthetic data needed.

\section{Finetuning Hyperparameters}
\label{sec:finetuning-hyperparams}
We performed a traditional hyperparameter sweep to select optimal hyperparameters for our Recover-LoRA method, traditional SFT LoRA baseline, as well as LLM QAT* baseline. 

For Recover-LoRA and SFT LoRA we utilized a learning rate of 5e-4, LoRA rank size of 64, LoRA alpha of 64, batch size of 1 with gradient accumulation of 32, a linear scheduler with 80 warm-up steps. Specifically for Recover-LoRA, we trained for 3 epochs, and for SFT LoRA we trained for 24k steps. 

For LLM QAT* we utilized a learning rate of 2e-5, a batch size of 1 with gradient accumulation of 32, a linear scheduler with 80 warmup steps and trained for 3 epochs. 

\section{Evaluation Model Details}
\label{sec:eval-models}
We evaluated Recover-LoRA on four different models, including Gemma2 2B \cite{team2024gemma}, Llama3.2 1B \cite{grattafiori2024llama}, DeepSeek-R1-Distill-Qwen 1.5B \cite{guo2025deepseek} and AMD-Olmo-SFT 1B \footnote{\label{note1} \url{https://huggingface.co/amd/AMD-OLMo-1B-SFT}}. These models were strategically chosen, given their different attention mechanisms: group-query attention (GQA) \cite{ainslie2023gqa}, and multi-head attention (MHA) \cite{chaudhari2021attentive}. 
In MHA, each head has its independent query, key and value projections, whereas in GQA, designated groups share the key and value projections. The AMD-OLMO-SFT 1B employs MHA, whereas Gemma2 2B model, Llama3.2 1B and DeepSeek-R1-Distill-Qwen 1.5B employ GQA. 

\end{document}